\documentclass[10pt,twocolumn,letterpaper]{article}

\usepackage{cvpr}      
\definecolor{cvprblue}{rgb}{0.21,0.49,0.74}
\usepackage[pagebackref,breaklinks,colorlinks,allcolors=cvprblue]{hyperref}
\usepackage{algorithm}
\usepackage{algpseudocode}
\usepackage{multirow}
\usepackage{url}

\def\paperID{7603} 
\def\confName{CVPR}
\def\confYear{2026}

\title{MoGAN: Improving Motion Quality in Video Diffusion via \\ Few-Step Motion Adversarial Post-Training}

\author{
\begin{tabular}{c}
Haotian Xue$^{1,2}$\thanks{Work done during internship at Adobe.} \quad
Qi Chen$^{1}$\thanks{Corresponding author.} \quad
Zhonghao Wang$^{1}$ \\
Xun Huang$^{1}$ \quad
Eli Shechtman$^{1}$ \quad
Jinrong Xie$^{1}$ \quad
Yongxin Chen$^{2}$
\end{tabular}
\\[6pt]
$^{1}$Adobe \qquad
$^{2}$Georgia Tech
}

\begin{document}
\maketitle

\begin{abstract}

Video diffusion models achieve strong frame-level fidelity but still struggle with motion coherence, dynamics and realism, often producing jitter, ghosting, or implausible dynamics. A key limitation is that the standard denoising MSE objective provides no direct supervision on temporal consistency, allowing models to achieve low loss while still generating poor motion.
We propose MoGAN, a motion-centric post-training framework that improves motion realism without reward models or human preference data. Built atop a 3-step distilled video diffusion model, we train a DiT-based optical-flow discriminator to differentiate real from generated motion, combined with a distribution-matching regularizer to preserve visual fidelity.
With experiments on Wan2.1-T2V-1.3B, MoGAN substantially improves motion quality across benchmarks. On VBench, MoGAN boosts motion score by +7.3\% over the 50-step teacher and +13.3\% over the 3-step DMD model. On VideoJAM-Bench, MoGAN improves motion score by +7.4\% over the teacher and +8.8\% over DMD, while maintaining comparable or even better aesthetic and image-quality scores. A human study further confirms that MoGAN is preferred for motion quality (52\% vs. 38\% for the teacher; 56\% vs. 29\% for DMD). Overall, MoGAN delivers significantly more realistic motion without sacrificing visual fidelity or efficiency, offering a practical path toward fast, high-quality video generation. Project webpage is: \url{https://xavihart.github.io/mogan/}.
\end{abstract}


\vspace{-5pt}
\section{Introduction}
\label{sec:intro}
\begin{figure}
    \centering
    \includegraphics[width=1\linewidth]{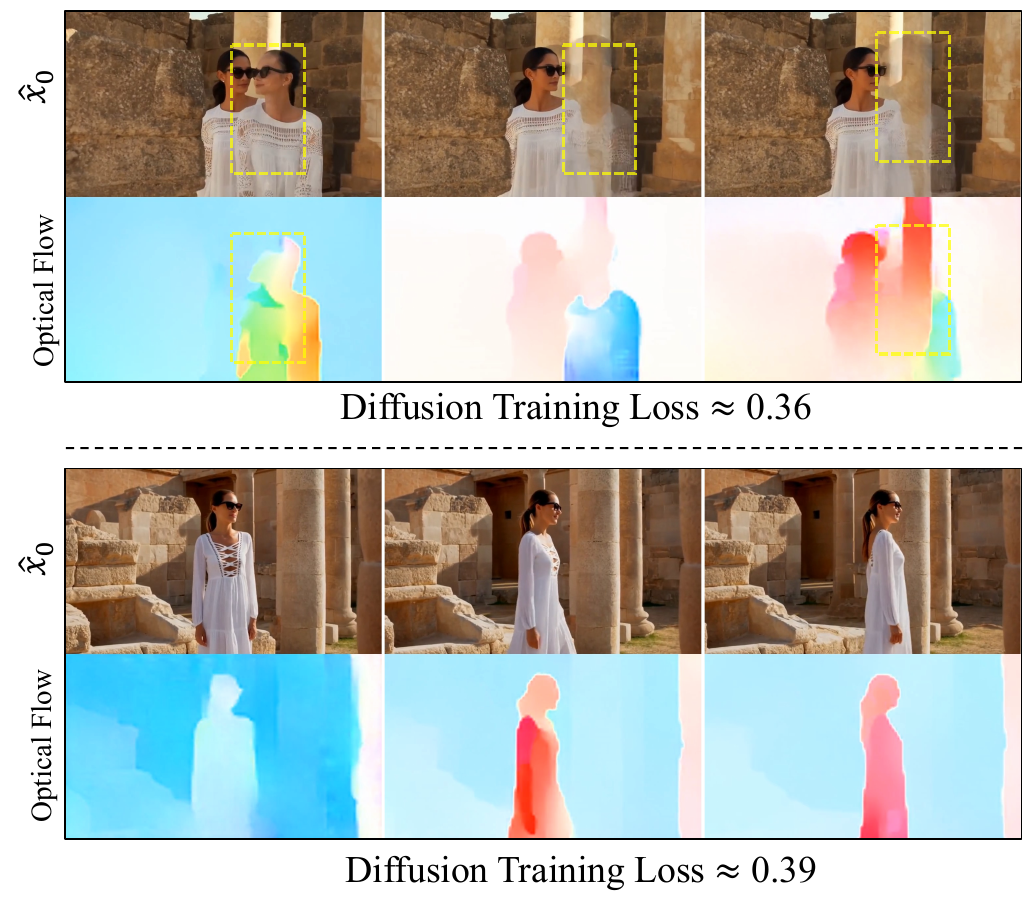}
    \vspace{-8pt}
    \caption{\textbf{Lower diffusion loss does not imply better motion.} Generated by the same model with different random seeds, the \emph{top block} achieves a lower diffusion training loss (\(\approx 0.36\)) but the predicted \(\hat{\mathbf{x}}_{0}\) exhibits ghosting, jitter, and incoherent optical flow in the highlighted regions. In contrast, the \emph{bottom block} has a slightly higher loss (\(\approx 0.39\)) yet produces smoother, more coherent motion with consistent flow fields. This discrepancy shows that pixelwise diffusion objectives (MSE) systematically under-penalize temporal artifacts and do not adequately optimize motion quality.}
    \label{teaser}
    \vspace{-10pt}
\end{figure}

Video diffusion models~\cite{wan2025wan,wan21github,yang2024cogvideox,kong2024hunyuanvideo,openai2024sora} can synthesize high fidelity frames, but they still struggle to model realistic temporal dynamics, which often yields motion that looks unrealistic. As illustrated in Figure~\ref{teaser}, a major cause is that the standard diffusion objective focuses only on per-frame pixel reconstruction, providing no explicit supervision for temporal consistency or motion realism. As a result, models may achieve low diffusion loss while still producing unstable motion. 

Prior work approaches this issue in three ways. Some methods inject external motion priors e.g., physics-simulation guidance or human-specified trajectories to steer generation~\cite{liu2024physgen,xue2025phyt2v,geng2025motion,burgert2025go}, but this reduces flexibility for open-ended prompts. Other approaches reshape the objective with motion-aware losses, such as applying optical-flow denoising as co-training, or align model to motion-sensitive representations~\cite{chefer2025videojam,bhowmik2025moalign}, but it still needs heavy computation to fine-tune since it introduces a new generation pipeline~\cite{chefer2025videojam}. Reinforcement-learning post-training (e.g., DPO/GRPO-style methods) has also been applied to video diffusion models~\cite{dpo,guo2025deepseek,wan21github,wu2025rewarddance,flow-grpo}, but existing approaches rely on vision-language reward models that evaluate only a small number of sampled frames (e.g., 8-frame clips~\cite{bai2025qwen25vl}).
Such reward signals capture semantics and aesthetics but do not accurately measure motion coherence, temporal consistency, or physical dynamics. In practice these objectives tend to modulate style (color, tone, theme) more than structure and fine-grained dynamics.


In this paper, we propose MoGAN (Motion-GAN Post-training), a novel post-training strategy with Motion-GAN that improves motion realism in video diffusion models by learning motion statistics directly from data. Rather than hand-designing an explicit motion loss, we introduce an adversarial objective in dense optical flow space. A frozen flow estimator extracts per frame motion fields from both real videos and model outputs. A Diffusion Transformer (DiT) based discriminator receives only the flow sequence and learns to distinguish real motion from generated motion.
We build Motion-GAN on a 3-step distilled video diffusion model, whose clean and efficient intermediate predictions allow reliable optical-flow extraction. We fine tune  video diffusion model initialized from DMD distilled Wan2.1-T2V-1.3B~\cite{wan2025wan}, and optimize two complementary objectives: a Motion-GAN loss in flow space and a distribution matching loss~\cite{yin2023dmd,yin2024dmd2} as regularization. The motion loss teaches realistic dynamics from data, while the distribution term preserves appearance fidelity and text alignment.  Importantly, inference remains unchanged, since our method keeps the efficient 3-step sampling path.

We evaluate our approach on VideoJAM-Bench~\cite{chefer2025videojam} and VBench~\cite{huang2024vbench} using both automatic metrics and controlled human preference studies. Across motion-related dimensions, including coherence, dynamics, and physical realism, our method is consistently preferred over the 3-step DMD baseline and achieves competitive or superior motion quality compared to the 50-step Wan2.1 model, while running over an order of magnitude faster. Human evaluations also indicate that our visual quality surpasses both the full-step and DMD baselines, demonstrating that improving motion does not come at the cost of appearance fidelity. As expected for few-step distilled models, text alignment is slightly weaker than the full-step generator but remains comparable to DMD. Ablation studies further validate the impact of each design choice. Overall, these results show that adversarial learning in flow space provides a strong and dense motion signal for video diffusion models, offering an effective alternative to RL-based post-training.

Our contribution can be summarized as follows:
\begin{itemize}
    \item We propose MoGAN, a new post-training framework using a DiT motion discriminator operating on optical-flow sequences.
    \item We introduce stabilizing techniques, including DMD regularization and discriminator regularization, to preserve fidelity and ensure robust training.
    \item Our 3-step model improves motion quality over both the 50-step Wan-2.1 baseline and the 3-step DMD baseline, achieving on average a 7.5\% gain over Wan-2.1 and a 10.5\% gain over Wan-DMD across VBench and VideoJAM-Bench.
\end{itemize}

\section{Related Works}\label{section:related works}

\begin{figure*}
    \centering
    \includegraphics[width=1\linewidth]{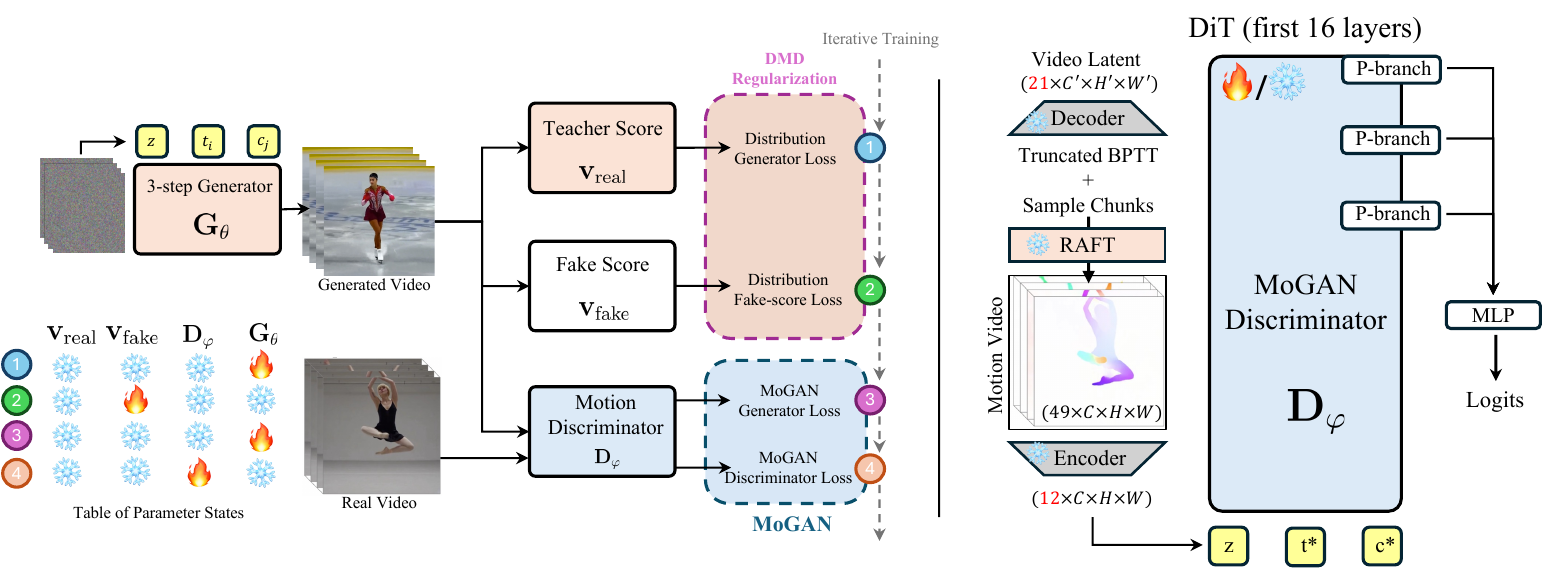}
    \caption{\textbf{Pipeline of the Proposed Few-Step Motion-GAN Post-Training}. Training loop iteratively optimizes four losses: two distribution-matching losses that regularize the student to remain close to the teacher distribution, and two MoGAN losses that directly improve motion quality. \textbf{(Left panel)}: given $t_i\in\{t_1,t_2,t_3\}$ and a condition $c_j$ from the prompt list, the few-step generator $\mathbf{G}_{\theta}$ produces an $x_0$ prediction. The teacher head $\mathbf{v}_{\text{real}}$ is frozen, while the student head $\mathbf{v}_{\text{fake}}$ learns to reflects the distribution modeled by $\mathbf{G}_{\theta}$. The \textbf{optical-flow centric} discriminator $\mathbf{D}_{\varphi}$ operates on dense optical-flows. \textbf{(Right panel)}: the DiT based optical flow discriminator, refer to Section~\ref{subsecion:design_of_motion_disc} for more details.}
    \label{fig:pipeline}
    \vspace{-8pt}
\end{figure*}

\paragraph{Post-Training for Diffusion Models}
Post-training typically falls into three families: (i) \textit{supervised fine-tuning (SFT)} on paired prompts or instructions to improve alignment and controllability~\cite{yang2024cogvideox,kong2024hunyuanvideo}; (ii) \textit{reinforcement learning with preferences or rewards}, including DPO/GRPO variants used in open and industrial systems~\cite{dpo,wan21github,wu2025rewarddance,flow-grpo,xue2025dancegrpo,guo2025deepseek}; and (iii) \textit{continuous reward feedback learning (ReFL)}, which optimizes differentiable scorers such as CLIP alignment, ImageReward/PickScore, or HPS v2 within the standard training loop~\cite{clark2024draft,xu2023imagereward,eyring2024reno,radford2021clip,kirstain2023pickapic,wu2023hpsv2}. While effective, RL-based methods often rely on undisclosed curation and careful reward design, and ReFL scorers can be gamed (reward hacking) and require costly, sometimes unstable backpropagation through long sampling chains~\cite{ho2020ddpm,karras2022edm}; these issues are amplified in video due to longer temporal horizons and the difficulty of defining scalable, informative motion rewards.

\paragraph{GAN-loss in diffusion post-training}
Adversarial objectives~\cite{goodfellow2014gan} (GAN-loss) have been coupled with diffusion training to preserve perceptual fidelity in few-step students: Adversarial Diffusion Distillation and its latent variant use a discriminator alongside score/distillation signals~\cite{add}. For video, Diffusion Adversarial Post-Training~\cite{lin2025seaweedAPT} adversarially fine-tunes one-step generators against real data. DMD-v2 finds that GAN-loss can help improve diversity in training~\cite{yin2024dmd2, xie2025turbofill}. These studies validate the utility of a discriminator for realism but mainly operate in pixel or latent space. \cite{optical-flow-guided} trains a GAN in optical-flow space using a single flow image and optimizes only the prompt embedding rather than the generator; consequently, it offers limited supervision for long-horizon motion and cannot learn sequence-level temporal statistics.

\paragraph{Improve Motion Quality of Video Diffusion} Several strategies target improving temporal realism of video diffusion model. Control/guidance injects external motion signals e.g. physics priors or human trajectories to constrain generation~\cite{liu2024physgen,xue2025phyt2v,geng2025motion,burgert2025go}, but this requires extra inputs/simulators and limits open-endedness. Objective design adds motion-aware losses, e.g., optical-flow denoising~\cite{chefer2025videojam} or alignment to motion-sensitive representations~\cite{bhowmik2025moalign}, which reduce artifacts but depend on estimator/backbone quality and incur heavy training. RL-based post-training optimizes DPO/GRPO-style objectives with motion rewards from heuristics or model judges~\cite{wan21github,wu2025rewarddance}, but it suffers from challenges in obtaining a good reward model or paired data for video. VLM-based judging supplies pseudo rewards or selections~\cite{zhang2025think,wu2025rewarddance}, but current VLMs struggle with long, higher-FPS video. Also some training-free methods e.g. FlowMo~\cite{shaulov2025flowmo} and RefDrop~\cite{fan2024refdrop} are proposed to improve consistency from adjust sampling process to reduce flicker, but it turns out that these method improves smoothness at the cost of dynamics. In contrast, we learn motion statistics directly in dense optical-flow space with a discriminator over flow fields, avoiding extra inference inputs and noisy preferences while targeting temporal coherence without sacrificing appearance.

\section{Background Knowledge}
\subsection{Video Diffusion Models}
Video diffusion models~\cite{wan2025wan,wan21github,yang2024cogvideox,kong2024hunyuanvideo,openai2024sora} learn a transport from a simple prior to the conditional data distribution. Under \emph{flow matching} (FM), we fit a time-dependent velocity field \(v_\theta\) that follows a prescribed path between prior and data. Let \(\mathbf{x}\in\mathbb{R}^{T\times C\times H\times W}\) be a video and \(\mathbf{c}\) a conditioning prompt. Define $p_0(\mathbf{x})=\mathcal{N}(\mathbf{0},\mathbf{I})$ and $p_1(\mathbf{x}\mid\mathbf{c})=p_{\mathrm{data}}(\mathbf{x}\mid\mathbf{c})$. FM fixes a coupling with intermediate samples $\,\mathbf{x}_t=\alpha(t)\mathbf{x}_0+\beta(t)\mathbf{x}_1\,$ for $t\in[0,1]$, where $\mathbf{x}_0\sim p_0$ and $\mathbf{x}_1\sim p_1(\cdot\mid\mathbf{c})$, with boundary conditions $\alpha(0)=1,\ \beta(0)=0,\ \alpha(1)=0,\ \beta(1)=1$. The oracle velocity along this path is $\mathbf{u}(\mathbf{x}_t\mid \mathbf{x}_0,\mathbf{x}_1)=\dot\alpha(t)\mathbf{x}_0+\dot\beta(t)\mathbf{x}_1$. The conditional FM objective trains $\mathbf{v}_\theta$ to match this velocity (with $t\sim\mathcal{U}(0,1)$):

\begin{equation*}
\mathcal{L}_{\mathrm{FM}}(\theta)=\mathbb{E}_{\mathbf{x}_0,\mathbf{x}_1,\mathbf{c},\, t}\left\|\mathbf{v}_\theta(\mathbf{x}_t,t,\mathbf{c})-\dot\alpha(t)\mathbf{x}_0-\dot\beta(t)\mathbf{x}_1\right\|_2^2.
\end{equation*}

The optimized \(\mathbf{v}_{\theta^*}\) can then be used to generate samples by solving the initial-value ODE: \(\frac{d\mathbf{x}_t}{dt}=\mathbf{v}_{\theta^*}(\mathbf{x}_t,t,\mathbf{c})\) for \(t\in[0,1]\) with \(\mathbf{x}_0\sim p_0\); integrating to \(t=1\) yields \(\mathbf{x}_1\sim p_{\mathrm{data}}(\cdot\mid\mathbf{c})\) which are synthesized videos. The above annotation works in pixel space, normally, we use an encoder $\mathcal{E}$ and decoder  $\mathcal{D}$ to project the diffusion into the latent space with $\mathbf{}{z}=\mathcal{E}(\mathbf{x})$, and $\mathbf{x}=\mathcal{D}(\mathbf{z})$ is used to decode then latents back to pixel space.

\subsection{Optical Flow Estimation}
Optical flow~\cite{zach2007tvl1,horn1981,lucas1981,brox2004,revaud2015epicflow} estimates a dense motion field \(\mathbf{o}(x,y)=(u(x,y),v(x,y))\) between two frames \(I_t\) and \(I_{t+1}\), typically under brightness constancy \(I_{t+1}(x+u,y+v)\approx I_t(x,y)\). Recent estimators leverage deep networks~\cite{dosovitskiy2015flownet,ilg2017flownet2,ranjan2017spynet,sun2018pwc} to learn optical flow from data. We adopt \emph{RAFT}~\cite{teed2020raft}, which extracts per-pixel features \(\phi_1,\phi_2\), builds an all-pairs 4D correlation volume \(\mathbb{C}\), and iteratively refines a flow estimate via \(\mathbf{o}^{(k+1)}=\mathbf{o}^{(k)}+\mathcal{F}\!\big(\phi_1,\phi_2,\mathbb{C},\mathbf{o}^{(k)}\big)\), yielding accurate displacement fields. In our setting, we denote the end-to-end optical flow estimator by \(\mathcal{F}\): given a video \(\mathbf{x}\in\mathbb{R}^{T\times C\times H\times W}\), it outputs an optical-flow sequence \(\mathbf{o}\in\mathbb{R}^{(T-1)\times 2\times H\times W}\), where \(\mathbf{o}[t]\) is the flow from frame \(\mathbf{x}[t]\) to \(\mathbf{x}[t+1]\). 

Optical flow provides a low-level, motion-centric representation that abstracts away appearance, whereas standard video diffusion models emphasize pixel reconstruction and underweight temporal dynamics. Introducing a motion-aware inductive bias e.g., via optical-flow-based objectives or discriminators has been shown to markedly improve motion quality~\cite{chefer2025videojam, optical-flow-guided}.

\section{Methods: Post-training with MoGAN } 
We introduce MoGAN, a scalable post-training framework for video diffusion models using an optical-flow-based motion discriminator. Some key design of our method includes: a scalable motion discriminator, a few-step video generation to enable reliable optical flow estimation, and finally training strategies to stablilize adversarial training e.g. some hyper parameters and some regularization terms. The idea is straightforward: use a motion-focused GAN to provide a continuous adversarial signal to enhance the motion quality of a few-step video diffusion model.

\subsection{Motion-Centric GAN combined with Distribution Matching Distillation}

We start our post-training from a warmed-up few-step video diffusion model $\mathbf{G}_{\theta}$ that can already generated clear intermediate samples. The few-step generator denoises the intermediate noisy samples as $\mathbf{G}_{\theta}(\mathbf{z}_{t_i},t_i,\mathbf{c})$, where $t_i \in \{t_1, t_2, t_3\}$ correspond to the distilled timesteps. We omit text condition $\mathbf{c}$ in the following  for simplicity. These discrete timesteps are distinct from the diffusion timesteps $t \in [0,1]$ used in the teacher model which are typically 1000-step.
Under the distribution matching distillation (DMD~\cite{yin2023dmd, yin2024dmd2}) settings, we also have a freezed teacher generator $\mathbf{v}_{\text{real}}$ and a fake score estimator $\mathbf{v}_{\text{fake}}$. The gradient of $\theta$ over the distribution matching loss $\mathcal{L}_{\mathrm{DMD}}=\mathbb{E}_{t}[\text{D}_{\text{KL}}(p_t^{\textbf{real}}||p_t^{\textbf{gen}})]$, where $p_t^{\textbf{real}}$ is the teacher distribution and $p_t^{\textbf{gen}}$ is the current generator distribution molded by $\mathbf{G}_{\theta}$. The update of DMD generator loss follows:

\begin{align}
\nabla_{\theta}\mathcal{L}_{\mathrm{DMD}}
\approx
\mathbb{E}_{t,\hat{\mathbf{z}}_0,t_i}\bigg[
& \left(
\mathbf{s}_{\mathrm{real}}\big(\mathbf{G}_{\theta}(\hat{\mathbf{z}}_{t_i}),t\big) \right. \\
& \left. - \mathbf{s}_{\mathrm{fake}}\big(\mathbf{G}_{\theta}(\hat{\mathbf{z}}_{t_i}),t\big)
\right)^{\top}
\frac{\partial \hat{\mathbf{z}}_{t_i}}{\partial \theta}
\bigg].\label{dmd_g_loss}
\end{align}

$\mathbf{s}_{\mathrm{real}}, \mathbf{s}_{\mathrm{fake}}$ are score for real and fake distribution, $\hat{\mathbf{z}}_{t_i}$ is noisy $\hat{\mathbf{z}}_0$ at timestep $t_i\in\{t_1, t_2, t_3\}$. We use the FM parameterization for all networks, but the  $\mathbf{s}, \mathbf{v}, \mathbf{z}_0$ predictions can be easily transformed from one to another. $\mathbf{s}_{\mathrm{fake}}$ also needs to be updated to match $\mathbf{G}_{\theta}$ each time when $\theta$ is updated, here we use flow matching loss to update it.

\begin{align}\label{dmd_c_loss}
\mathcal{L}_{\mathrm{fake}}^{\phi}=\mathbb{E}_{\mathbf{z}_0,\mathbf{z}_1,\mathbf{c},\, t}\left\|\mathbf{v}^{\phi}_{\mathrm{fake}}(\mathbf{z}_t,t)-\dot\alpha(t)\mathbf{z}_0-\dot\beta(t)\mathbf{z}_1\right\|_2^2.
\end{align}

The DMD generator loss \textbf{stabilizes} training by imposing \textbf{distributional regularization} on the generator updates. 

Then we introduce a Generative Adversarial objective that focuses on \emph{motion} by operating in pixel space optical-flow. Let \(\mathcal{F}\) be a frozen, differentiable flow estimator that maps a video \(\mathbf{x}\in\mathbb{R}^{T\times C\times H\times W}\) to a dense flow stack \(\mathbf{o}=\mathcal{F}(\mathbf{x})\in\mathbb{R}^{(T-1)\times 2\times H\times W}\). For each distilled timestep \(t_i\), we obtain a decoded clip \(\mathbf{x}^{\mathrm{gen}}_\theta=\mathcal{D}\!\left(\mathbf{G}_\theta(\hat{\mathbf{z}}_{t_i},t)\right)\) and a real clip \(\mathbf{x}^{\mathrm{real}}\) from data; their flows are \(\mathbf{o}^{\mathrm{gen}}_\theta=\mathcal{F}(\mathbf{x}^{\mathrm{gen}}_\theta)\) and \(\mathbf{o}^{\mathrm{real}}=\mathcal{F}(\mathbf{x}^{\mathrm{real}})\). A motion discriminator \(\mathbf{D}_\varphi\) consumes flow and outputs a real value (Section~\ref{subsecion:design_of_motion_disc} will introduce some details of \(\mathbf{D}_\varphi\)). We adopt the \textbf{logistic  GAN} loss~\cite{goodfellow2014gan}, including GAN discriminator loss $\mathcal{L}_{\mathrm{GAN}}^{\theta}$, and GAN generator loss $\mathcal{L}_{\mathrm{GAN}}^{\varphi}$ for motion GAN, which is defined as:

\begin{align}
\mathcal{L}_{\mathrm{GAN}}^{\varphi}
&= \mathbb{E}_{t,\mathbf{c}}
   \big[\operatorname{g}\!\big(-\mathbf{D}_{\varphi}(\mathbf{o}^{\mathrm{real}})\big)
   + \operatorname{g}\!\big(\mathbf{D}_{\varphi}(\mathbf{o}^{\mathrm{gen}}_\theta)\big)\big],
   \label{eq:motiongan_d} \\[4pt]
\mathcal{L}_{\mathrm{GAN}}^{\theta}
&= \mathbb{E}_{t,\mathbf{c}}
   \big[\operatorname{g}\!\big(-\mathbf{D}_{\varphi}(\mathbf{o}^{\mathrm{gen}}_\theta)\big)\big].
   \label{eq:motiongan_g}
\end{align}

where $\operatorname{g}(x) = \log\!\big(1 + e^{x}\big)$. We apply R1 and R2 regularization~\cite{roth2017stabilizing} on the discriminator to prevent overfitting and stabilize adversarial training. These regularizers have been shown to be crucial for maintaining training stability when optimizing with the GAN objective (see Ablation Study), and are formally defined as:

\begin{align}
\mathcal{L}_{\mathrm{R1}}^{\varphi}
&= \left\| 
\mathbf{D}_{\varphi}(\mathbf{o}^{\mathrm{real}})
- \mathbf{D}_{\varphi}\!\big(\mathcal{N}(\mathbf{o}^{\mathrm{real}},\,\sigma \mathbf{I})\big)
\right\|_2^{2},
\label{eq:r1_loss} \\[4pt]
\mathcal{L}_{\mathrm{R2}}^{\varphi}
&= \left\| 
\mathbf{D}_{\varphi}(\mathbf{o}^{\mathrm{gen}}_{\theta})
- \mathbf{D}_{\varphi}\!\big(\mathcal{N}(\mathbf{o}^{\mathrm{gen}}_{\theta},\,\sigma \mathbf{I})\big)
\right\|_2^{2}.
\label{eq:r2_loss}
\end{align}

The final optimization loss for Motion-GAN update be written as the follows, with $\lambda_{1}$ and $\lambda_{2}$ to control weights for Motion-GAN updates, and $\lambda_{\mathrm{R1}}$ and $\lambda_{\mathrm{R2}}$ for regularization:

\begin{equation}
\mathcal{L}_{\mathrm{GAN}}
=
\underbrace{
\lambda_{1}\,\mathcal{L}_{\mathrm{GAN}}^{\theta}
}_{\text{Generator Loss}}
+
\underbrace{
\lambda_{2}\,\mathcal{L}_{\mathrm{GAN}}^{\varphi}
+ \lambda_{\mathrm{R1}}\,\mathcal{L}_{\mathrm{R1}}^{\varphi}
+ \lambda_{\mathrm{R2}}\,\mathcal{L}_{\mathrm{R2}}^{\varphi}
}_{\text{Discriminator Loss}}.
\end{equation}

\subsection{Training Strategy}
We first warm up the few-step generator so that it can produce sufficiently clean intermediate predictions, ensuring that the subsequent optical-flow estimates are meaningful. We then alternate updates between the DMD regularization loss and the proposed Motion-GAN (MoGAN) loss, the training loop is also illustrated in the left panel of Figure~\ref{fig:pipeline}. We combine R1 and R2 as part of MoGAN discriminator loss to regularize the motion discriminator. We update DMD critic more frequently following~\cite{causvid}. More details about training hyper-parameters are put in the experiment section.

\begin{algorithm}[t]
\caption{\textbf{Motion GAN Post-Training}}
\label{alg:motiongan}
\begin{algorithmic}[1]
\Require 
Pretrained 3-step video generator $\mathbf{G}_{\theta}$ with distilled timstep $t_1,t_2,t_3$; 
real video dataset $\{\mathbf{x}^{\mathrm{real}}_k\}$; 
prompt set $\{\mathbf{c}_j\}$; 
flow estimator $\mathcal{F}$; 
motion discriminator $\mathbf{D}_{\varphi}$.

\For{training loop}
    \State $t\sim [0,1]$, $\mathbf{c}\sim\{\mathbf{c}_j\}$
    \State Backward simulation~\cite{yin2024dmd2} with $\mathbf{G}_{\theta}$ to get $\hat z_{0}$
    \State $t_i\in\{t_1, t_2, t_3\}$
    \State $\hat{\mathbf{z}}_{t_i}=(1-t_i)*\hat{\mathbf{z}}_{t_0}+t_i*\mathcal{N}(0, \textbf{I)}$
    \State Update few-step generator $\theta$ with Eq~\ref{dmd_g_loss}.
    \For{fake score training loop}
          \State Backward simulation~\cite{yin2024dmd2} to get $\hat{\mathbf{z}}_{0}$
          \State $\mathbf{z}_0=\hat{\mathbf{z}}_{0}$
          \State $\mathbf{z}_1\sim\mathcal{N}(0, \textbf{I)}$
          \State Update fake score $\phi$ with Eq~\ref{dmd_c_loss}.
    \EndFor
    \State $\mathbf{x}^{\mathrm{gen}}_\theta=\mathcal{D}\!\left(\mathbf{G}_\theta(\hat{\mathbf{z}}_{t_i},t,\mathbf{c})\right)$
    \State $\mathbf{x}^{\mathrm{real}}\sim\{\mathbf{x}^{\mathrm{real}}_k\}$
    \State $\mathbf{o}^{\mathrm{gen}}_\theta=\mathcal{F}(\mathbf{x}^{\mathrm{gen}}_\theta)$ and $\mathbf{o}^{\mathrm{real}}=\mathcal{F}(\mathbf{x}^{\mathrm{real}})$
    \State Update motion discriminator $\varphi$ with
    Eq~\ref{eq:motiongan_d}.
     
    \State Regularize $\varphi$ with Eq~\ref{eq:r1_loss}. and  ~\ref{eq:r2_loss}.
    \State Update  $\theta$ with motion-GAN generator loss in Eq~\ref{eq:motiongan_g}.
\EndFor
\end{algorithmic}
\end{algorithm}


\subsection{DiT-based Motion Discriminator}\label{subsecion:design_of_motion_disc}
We illustrate the model structure of $\mathbf{D}_{\varphi}$ in right panel of  Figure~\ref{fig:pipeline}. Assume we have to optimize our video generator which is parameterized by $\mathbf{G}_\theta$, one imporant part of applying adversarial post-training is to use a discriminator $\mathbf{D}_{\varphi}$. 

Given RAFT-predicted optical flow~\cite{teed2020raft} for each frame of a generated video, we append one additional optical flow map to the last to make it the same size of the input video. Then we compute the per-pixel flow magnitude, and stack this magnitude with the original flow channels to form a three-channel motion representation. This motion tensor is fed to a motion discriminator adapted from the Wan2.1 DiT backbone~\cite{Peebles_2023_ICCV,wan2025wan}: following~\cite{lin2025seaweedAPT}, we add lightweight multi-scale heads at several depths, each using cross-attention with an auxiliary token followed by a small MLP (P-Branch in Figure~\ref{fig:pipeline} (right)); their outputs are concatenated and passed to a final MLP to produce a single scalar logit. We fix the diffusion timestep and use  “a video with good motion” as prompt $\mathbf{c}^*$, so that the discriminator focuses purely on motion realism, and we fix $t^*=0$ to make it deterministic.

Naively computing optical flow requires fully decoding all latents through the chunk-recurrent Wan decoder~\cite{wan2025wan}, which is slow and memory-intensive. To make this practical, we combine truncated backpropagation through time (BPTT)~\cite{werbos1990bptt}, gradient checkpointing, and chunk subsampling/early stopping: we decode only a contiguous subset of chunks, compute flows and adversarial losses within this window, detach the recurrent state at its boundary, and terminate decoding once the window is covered. This keeps memory bounded, reduces redundant decoding, and still provides informative gradients for the motion-aware GAN objective. We put more details about the design of the MoGAN discriminator in the Appendix.

\section{Experimental Setups}
\label{sec:experiments}

\subsection{Settings of MoGAN Post-Training}
We fine-tune our model from the Wan2.1-T2V-1.3B model~\cite{wan21github} as our base generator. 
The optimization uses AdamW with a learning rate of \(1\times10^{-5}\). 
We set the Motion-GAN loss weights to \(\lambda_{1}=\lambda_{2}=0.5\) and the regularization coefficients to \(\lambda_{\mathrm{R1}}=\lambda_{\mathrm{R2}}=0.3\), 
with a noise perturbation of \(\sigma=0.01\). 
The discriminator is trained with a batch size of \(64\), while the generator uses a smaller batch size of \(16\) due to higher memory consumption. 
Training is conducted for \(800\) steps. 
We distill using a prompt set \(\{c_j\}\) containing \(5\text{K}\) diverse text prompts, 
and employ a real motion dataset \(\{x_k^{\mathrm{real}}\}\) of \(15\text{K}\) videos curated for rich and dynamic motion content. 

For the MoGAN discriminator, we use 12 latent chunks corresponding to 49 frames in pixel space. For the DiT backbone of the dicriminator, we adopt the first 16 layers of the pretrained Wan2.1-T2V-1.3B DiT as the backbone and attach prediction branches at layers 7, 13, and 15.

\subsection{Baselines and Metrics}
We conduct both qualitative and quantitative experiments to demonstrate the superiority of our proposed Motion-GAN in enhancing motion quality. We evaluate on VBench~\cite{huang2024vbench} and VideoJAM-Bench~\cite{chefer2025videojam}. For motion quality assessment, we adopt the motion smoothness and dynamics degree metrics from VBench, while aesthetic and image quality scores are used to measure video quality. Following~\cite{chefer2025videojam}, we also compute a 
\textbf{motion score} defined as the mean of motion smoothness (based on frame interpolation) and dynamics degree (based on optical flow), which penalizes static videos and provides a fairer measure of overall motion quality. For each prompt we sample $5$ different seeds and use same seed across different models for fair comparison, and calculate  the mean across seeds for different metrics. 

To obtain a more comprehensive evaluation, we further conduct human studies comparing our method with baseline models in Section~\ref{subsection:human_survey}.

\begin{figure*}
    \centering
    \includegraphics[width=1\linewidth]{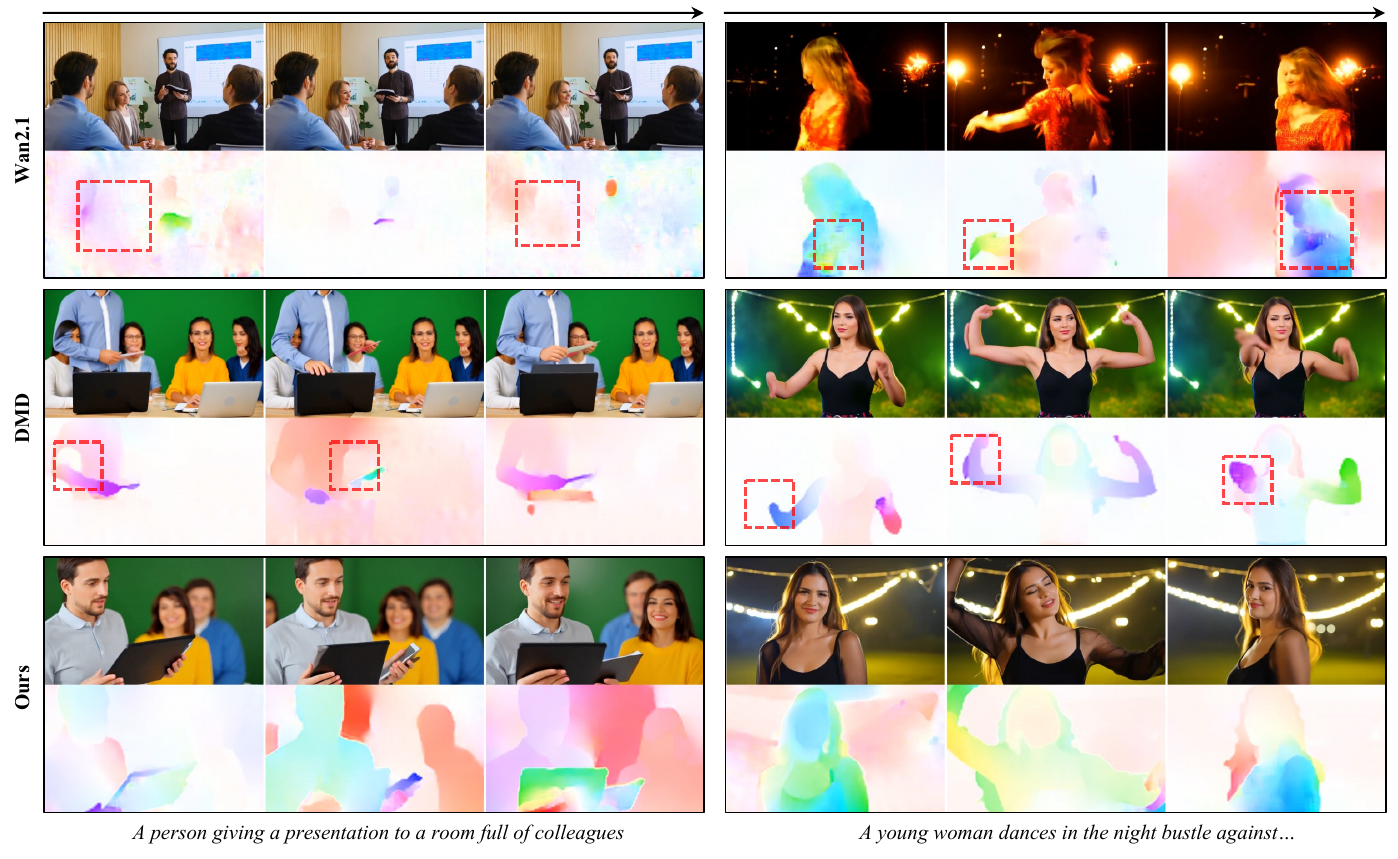}
    \caption{\textbf{Qualitative comparison across models}:
    For both prompts, we show video clips alongside the optical-flow visualization for three models: Wan2.1 (50-step), DMD-only (3-step), and our Motion-GAN post-trained model, all sampled with the \textbf{same seed}. Motion artifacts that are sometimes subtle in pixel space become clearly visible in the optical-flow maps and are highlighted with red boxes. More visualizations are in the Appendix.}
    \label{fig:vis_general}
\end{figure*}

\section{Results}


\begin{table*}[t]
\centering
\small
\resizebox{0.98\linewidth}{!}{
\setlength{\tabcolsep}{8pt}
\renewcommand{\arraystretch}{1.25}
\begin{tabular}{l|l|ccc|cc}
\hline
\textbf{Dataset} & \textbf{Model} & \textbf{Smoothness (\%)} & \textbf{Dynamics Degree } & \textbf{Motion Score} & \textbf{Aesthetic} & \textbf{Image Quality} \\
\hline\hline
\multirow{3}{*}{VBench~\cite{huang2024vbench}} 
& Wan \emph{[50-steps]} & 98.0 & 0.83 & 0.905 & 0.57 & 0.66 \\
& FlowMo \emph{[50-steps]}  & 98.6 & 0.82 $\downarrow$ & 0.903 & 0.58 & 0.64 \\
& DMD-only \emph{[3-steps]}  & \textbf{98.8} & 0.73 $\downarrow$  & 0.859 & 0.57 & \textbf{0.69} \\
& MoGAN (Ours) \emph{[3-steps]}  & 98.6 & \textbf{0.96} $\uparrow$  & \textbf{0.973} & \textbf{0.59} & 0.68 \\
\hline
\multirow{3}{*}{VideoJAM-Bench~\cite{chefer2025videojam}} 
& Wan \emph{[50-steps]} & 97.9 & 0.85 & 0.915 & 0.55 & 0.63 \\
& DMD-only \emph{[3-steps]}  & \textbf{98.5} & 0.81$\downarrow$ & 0.898 & \textbf{0.57} & 0.65 \\
& MoGAN (Ours) \emph{[3-steps]}  & \textbf{98.5} & \textbf{0.98}$\uparrow$& \textbf{0.983} & \textbf{0.57} & \textbf{0.66} \\
\hline
\end{tabular}
}
\caption{\textbf{Motion Quality Improvement of Few-Step Models.} 
We compare the motion and video quality of our 3-step models with the full 50-step Wan2.1 baseline on both VBench and VideoJAM-Bench. 
While DMD improves motion smoothness, it tends to {\color{red}\textbf{reduce dynamics}} , leading to \textbf{more static} videos. 
Our approach achieves a better trade-off, enhancing both \textbf{motion score} (average of smoothness and dynamics) and temporal realism, 
while maintaining comparable aesthetic and image quality.}
\label{tab:motion_quality_comparison}
\vspace{-0.3cm}
\end{table*}

\subsection{Numerical Results of Auto Metrics}
We present our quantitative results using metrics from VBench in Table~\ref{tab:motion_quality_comparison}, evaluating our method on the VBench and VideoJAM-Bench datasets. We compare our $3$-step Motion-GAN post-trained model against two key baselines: ($1$) the original Wan2.1-1.3B ($50$-step) generator, representing the non-distilled model, and ($2$) DMD-only ($3$-step), a distilled model trained only with distribution matching loss.

On V-Bench, we observe that the DMD-only baseline suffers a severe degradation in motion dynamics. It achieves high smoothness ($98.8\%$) but experiences a catastrophic \textbf{$12.0\%$ drop in Dynamics Degree} ($0.73$) compared to the $50$-step Wan2.1 model ($0.83$). This confirms it tends to produce more static videos, resulting in a poor overall Motion Score of $0.859$. In stark contrast, our $3$-step model not only maintains high smoothness ($98.6\%$) but dramatically \textbf{boosts the Dynamics Degree to $0.96$}, a $7.5\%$ \textit{increase} over the $50$-step baseline. This results in the highest \textbf{Motion Score ($0.973$)}, far surpassing both the original  and the DMD-only models. This superior performance is mirrored on the VideoJAM-Bench dataset, which includes much more difficult motion prompts than VBench. The DMD-only model again sacrifices motion ($0.81$ Dynamics Degree), leading to a low Motion Score ($0.830$). Our model, however, achieves the best-in-class \textbf{Dynamics Degree ($0.98$)} and the highest \textbf{Motion Score ($0.983$)}, representing a $7.4\%$ improvement in dynamics over the $50$-step baseline. We also take numbers from FlowMo~\cite{shaulov2025flowmo} paper on VBench and MoGAN achieves much better results than FlowMo.

Critically, these substantial gains in motion quality do not compromise perceptual quality. Our model consistently matches or slightly improves upon the \textbf{Aesthetic} and \textbf{Image Quality} scores of the original $50$-step model. We caution that VBench scores should be interpreted with care: several metrics are not sufficiently discriminative in our setting and can obscure meaningful differences between models. Accordingly, we complement these numbers with qualitative visualizations (Section~\ref{subsection:vis_results}) and a large-scale human evaluation (Section~\ref{subsection:human_survey}) to compare performance across models more reliably.


\begin{figure}
    \centering
    \includegraphics[width=1\linewidth]{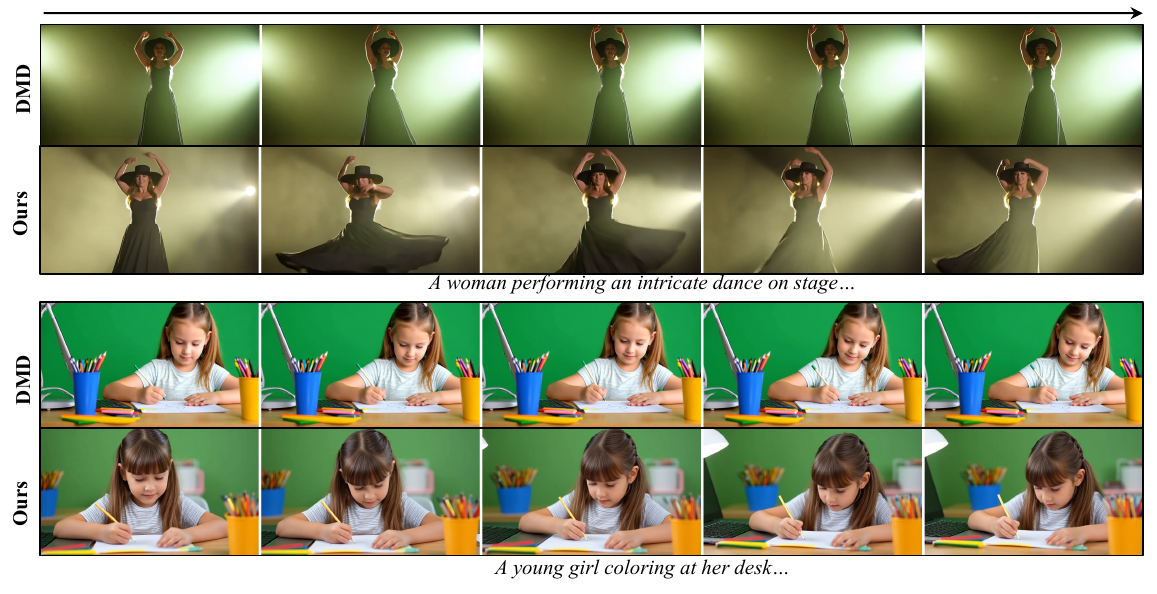}
    \caption{\textbf{Our Model Improves Smoothness Without Sacrificing Dynamics}: Motion-GAN post-training generates more realistic motion by balancing dynamics and smoothness. 
    In both examples, the DMD-distilled model tends to produce \textbf{overly static} videos, while our method generates smoother and more naturally dynamic motion.}
    \label{fig:vis_dynamics}
    \vspace{-8pt}
\end{figure}

\subsection{Visualization Results}\label{subsection:vis_results}

\textbf{Improved Video Motion Quality} Motion-GAN produces smoother and more natural motion than both Wan2.1 and DMD (Figs.~\ref{fig:vis_general},~\ref{fig:vis_dynamics}). In Fig.~\ref{fig:vis_general}, the \emph{“person giving a presentation”} scene illustrates how temporal artifacts in pixel space are revealed by optical flow: Wan2.1 exhibits background flickering that appears as noisy, inconsistent flow, while in the DMD baseline the woman’s head in the background suddenly appears and disappears, leading to discontinuous flow patterns. In the \emph{“woman dancing”} scene, Wan2.1 introduces background distortions that again yield noisy flow estimates, and DMD causes the dancer’s arms to warp, which is mirrored by distorted flow fields. In contrast, our Motion-GAN model produces stable and coherent optical flow across frames, indicating substantially improved motion quality.
\vspace{-0.3cm}
\paragraph{Restored Motion Dynamics} DMD boosts smoothness score than full-step Wan but generate much slower videos. e.g. In Fig.~\ref{fig:vis_dynamics} The dancer barely moves (first prompt) and the camera is always static (second prompt), producing smooth-but-static sequences; similar attenuation appears in lights and rope amplitude in Fig.~\ref{fig:vis_general}. By contrast, our post-trained model exhibits stronger dynamics while maintaining smooth motion. The balance arises from adversarial flow supervision that learns discrimination from motion rich videos in our dataset, which preserves motion amplitude rather than damping it.
\vspace{-0.3cm}
\paragraph{Balanced Color Aesthetics} DMD tends to over-saturate colors and push highlights (e.g., ocean, stage lights, green background), whereas Wan2.1 preserves tonal variety but suffers temporal artifacts (Fig.~\ref{fig:vis_general}). Motion-GAN lands between them, keeping natural skin tones, contrast, and shading while maintaining temporal coherence. The result is visually pleasing, less “neon,” and more consistent across frames, matching Wan2.1’s fidelity without reintroducing a lot of flicker or distortions. It may partially due to: over-saturation tends to generate  noisy optical flow, which will be punished by discriminator. It is also refelected in the human survey in the next section.

\begin{table}[t]
\centering
\small
\resizebox{1\linewidth}{!}{
\setlength{\tabcolsep}{8pt}
\renewcommand{\arraystretch}{1.25}
\begin{tabular}{l|ccc|cc}
\hline
\textbf{Model} & \textbf{Smoothness (\%)} & \textbf{Dynamics} & \textbf{Motion Score} & \textbf{Aesthetic} & \textbf{Image Quality} \\
\hline\hline
Ours & \textbf{98.5} & \textbf{0.98} & \textbf{0.983} & \textbf{0.57} & 0.66 \\
w/o DMD Loss & 99.1 & 0.35 & 0.674 & 0.50 & 0.55 \\
w/o R1 $\&$ R2  & 95.1 & 0.86 & 0.905 & 0.55 & 0.64 \\
w/o Optical Flow & 98.0 & 0.85 & 0.915 & 0.56 & \textbf{0.67} \\
\hline
\end{tabular}
}
\caption{\textbf{Auto-Metrics for Ablation Study.} We compare our full model to three variants: (i) without DMD co-training for distribution regularization, (ii) without R1/R2 regularization on the motion discriminator, and (iii) replacing the optical-flow-based GAN loss with a video-space GAN loss. Also refer to Figure~\ref{fig:ablation_1} to see visual quality. The experiments are run on VBench.}
\vspace{-0.5cm}
\end{table}\label{table:ablation}

\subsection{Human Evaluation with Survey}\label{subsection:human_survey}
\begin{figure}
    \centering
    \includegraphics[width=1\linewidth]{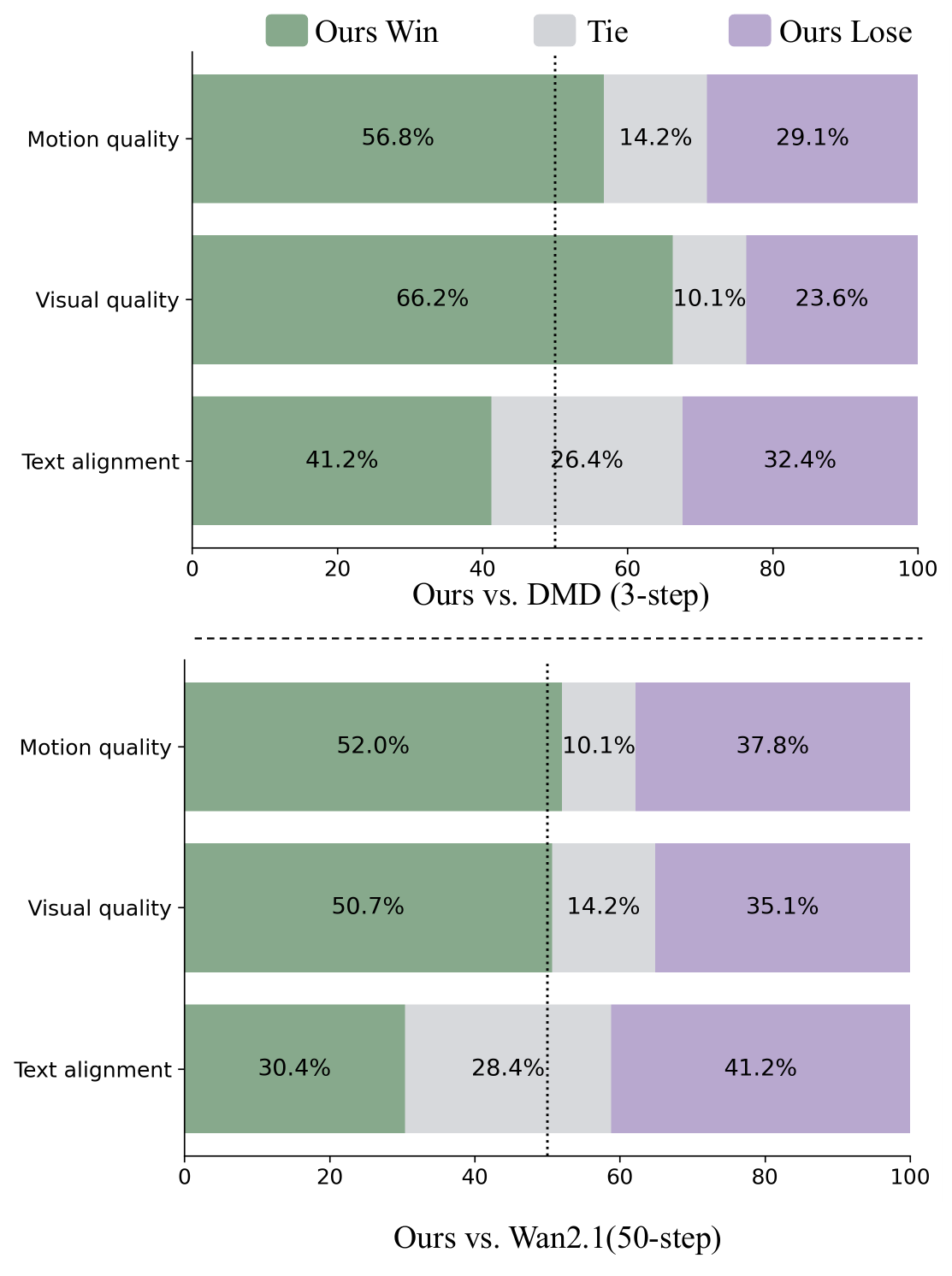}
    \caption{\textbf{Results of Human Survey.} Side-by-side human preference study comparing our 3-step Motion-GAN with DMD (3-step) and Wan2.1 (50-step).}
    \label{fig:ablation_1}
    \vspace{-15pt}
    
\end{figure}

We conducted a human evaluation over $148$ videos, each evaluated on three criteria: motion quality, visual quality, and text alignment. Each video received five independent annotations, resulting in a total of $2,220$ responses per comparison setting: (1) Ours vs. DMD (3-step) and (2) Ours vs. Wan2.1 (50-step). The win-tie-lose results are shown in Figure~\ref{subsection:human_survey}. Across both comparisons, our model consistently outperforms DMD and Wan2.1 in terms of motion quality and visual quality. For text alignment, our model performs slightly below the Wan2.1 50-step baseline, which is expected given that both DMD and our model operate in a 3-step distilled setting, inheriting some alignment limitations from the distilled backbone.


\subsection{Ablation Study}

\begin{figure}
    \centering
    \includegraphics[width=1\linewidth]{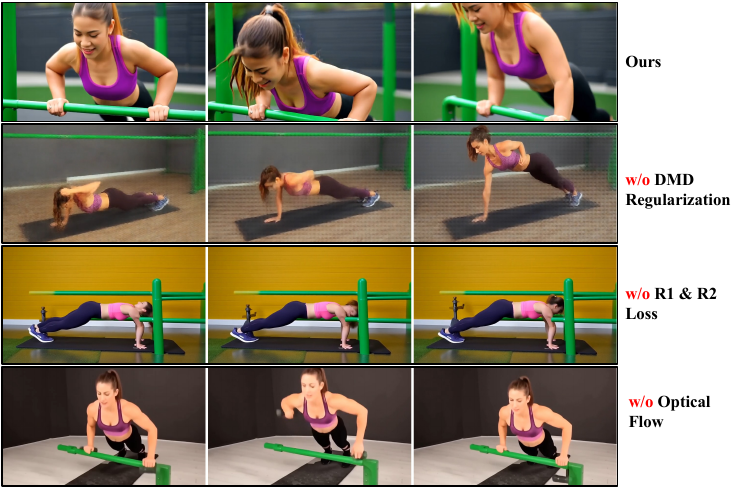}
    \caption{\textbf{Ablation Study Visualization}: Removing DMD regularization leads to training collapse, as the generator drifts away from the teacher distribution. Eliminating the R1/R2 regularization causes the discriminator to overpower the generator, hindering stable optimization. Removing the optical-flow will not bring motion improvements and results in blurrier frames. More results are in the Appendix.}
    \label{fig:ablation_1}
    \vspace{-0.3cm}
    \end{figure}


\paragraph{Remove DMD Regularization}
Removing the DMD tether lets the student drift from the teacher distribution and exploit the adversarial objective: micro-jitters and flow noise are amplified to “fool” the critic, yielding mode collapse and temporal artifacts in Fig.~\ref{fig:ablation_1}. From Table~\ref{table:ablation} we can also see that both motion and visual score are much worse compared with our models.
\vspace{-0.3cm}
\paragraph{Remove Discriminator Regularization}
Without R1/R2 regularization, the motion discriminator rapidly overfits and overpowers the generator. We observe that removing these terms leads to highly unstable training, characterized by sharp spikes in the GAN loss. As shown in Fig.~\ref{fig:ablation_1}, dropping R1/R2 yields noticeably worse videos: motion quality and visual quality degrade. This trend is consistent with the quantitative results in Table~\ref{table:ablation}, where motion smoothness decreases without discriminator regularization.
\vspace{-0.3cm}
\paragraph{Use Video-only Discriminator} To assess the benefit of computing the GAN loss in optical-flow space, we evaluate a variant that applies the adversarial objective directly in the video latent space, removing the flow estimator and downstream encoder/decoder. While this restores more natural color (close to undistilled Wan2.1), it retains DMD-like static motion (Table~\ref{table:ablation}) and it looks more blury (Fig.~\ref{fig:ablation_1}) than all other methods. More results can be referred to in the Appendix.


\section{Discussion}
Video diffusion models remain challenged by capturing realistic motion despite strong advancements in appearance fidelity. Our findings suggest that learning directly in flow space offers a powerful complementary signal to flow-matching training, helping few-step models acquire smoother and more coherent dynamics. By introducing a scalable flow-based discriminator and stabilizing training with DMD regularization, we observe consistent improvements across VBench, VideoJAM-Bench, and human preference studies.

However, the method inherits several limitations. First, it depends on pixel-space decoding and a frozen 2D optical-flow estimator, whose non-physical nature can misinterpret occlusions, out-of-plane motion, or fast articulation. Second, optical flow becomes unreliable for extremely small motions or complex depth changes. Future work may explore latent-space motion surrogates, 3D-consistent or geometry-aware motion fields, or hybrid physical motion priors to further enhance temporal realism.

\section{Conclusion}
We introduced MoGAN as a post-training framework that enhances motion quality in few-step video diffusion models by pairing a flow-space adversarial objective with distribution matching regularization. The approach preserves image fidelity, maintains inference speed, and improves temporal coherence and dynamics over both a 50-step baseline and a DMD-only distilled model. Our results suggest that adversarial learning in optical-flow space is a scalable and effective direction for building video generators with more realistic motion.



{
    \small
    \bibliographystyle{ieeenat_fullname}
    \bibliography{main}
}

\newpage
\appendix

Please refer to our project webpage at \url{https://xavihart.github.io/mogan} for more video visualizations of this paper.

\section{Details about MoGAN Discriminator}
\paragraph{Input} Given the raw pixel-space optical flow $\mathbf{o_{\text{raw}}}\in\mathbb{R}^{(T-1)\times 2\times H\times W}$ of a generated video predicted by \emph{RAFT}~\cite{teed2020raft}, we first append one frame by duplicating the last flow field so the stack has $T$ frames: $\tilde{\mathbf{o}}\in\mathbb{R}^{T\times 2\times H\times W}$. We then compute the per-pixel flow magnitude $m=\|\tilde{\mathbf{o}}\|_2$ (the $\ell_2$ norm over the two flow channels) and concatenate it as an additional channel, yielding a three-channel flow tensor $\mathbf{o}\in\mathbb{R}^{T\times 3\times H\times W}$ that matches the video input shape (e.g., $81\times3\times480\times832$ for 480p in Wan2.1-T2V). Finally, we feed $\mathbf{o}$ to a motion discriminator $\mathbf{D}_{\varphi}$ adapted from a scalable Diffusion Transformer (DiT)~\cite{Peebles_2023_ICCV}.  For the remaining DiT inputs, we fix the diffusion timestep to $t^\ast=0$ and set the condition token $c^\ast$ to the prompt embedding of “a video with good motion.”

\paragraph{Prediction Head} Following \cite{lin2025seaweedAPT}, we attach lightweight prediction heads at three fixed depths of the pre-trained Wan2.1 DiT to capture multi-scale features. Each head performs cross-attention with an auxiliary token and then applies a small MLP. The head outputs are concatenated and fed to a final MLP, producing a single scalar logit.

\paragraph{Save Memory for Backpropgation} One big problem is the efficiency of obtaining $\mathbf{o}_{\text{raw}}$, since we need to decode the latent $z$ into pixel space first. Also, the Wan decoder~\cite{wan2025wan} is chunk-recurrent (RNN-like), decoding each latent chunk into a short frame segment while propagating a hidden state, so fully unrolling it to recover all frames is slow and prone to OOM. We address this by combining Truncated Back Propagation Through Time (BPTT~\cite{werbos1990bptt}) with gradient checkpointing and chunk subsampling/early stopping: we sample $L$ continuous chunks out of a total of $K$ chunks, unroll only a window of these chunks with gradients (e.g., $L{=}12$ of $K{=}21$), compute flows within this window, detach the hidden state at the boundary, and terminate decoding once the $L$-th chunk is reached. This keeps memory bounded, avoids redundant decoding, and preserves informative gradients for the adversarial signal.

\section{Training Details}
We conduct experiments on 16 H200 GPUs with 141G memory. The model is implemented in PyTorch, and we apply FSDP and gradient checkpointing to save the memory. Training parameters are same as that listed in the main paper: the optimization uses AdamW with a learning rate of \(1\times10^{-5}\). 
We set the Motion-GAN loss weights to \(\lambda_{1}=\lambda_{2}=0.5\) and the regularization coefficients to \(\lambda_{\mathrm{R1}}=\lambda_{\mathrm{R2}}=0.3\), 
with a noise perturbation of \(\sigma=0.01\).

We iteratively train the four losses, the order (1) Update few-step generator with DMD generator loss, (2) Update fake score for $4$ iterations over DMD critic loss, (3) Update generator with MoGAN generator loss, (4) Update MoGAN discriminator with MoGAN discriminator loss. We do early stop at step $800$ as our final checkpoint.

\section{Human Survey Details}
\begin{figure}
    \centering
    \includegraphics[width=0.9\linewidth]{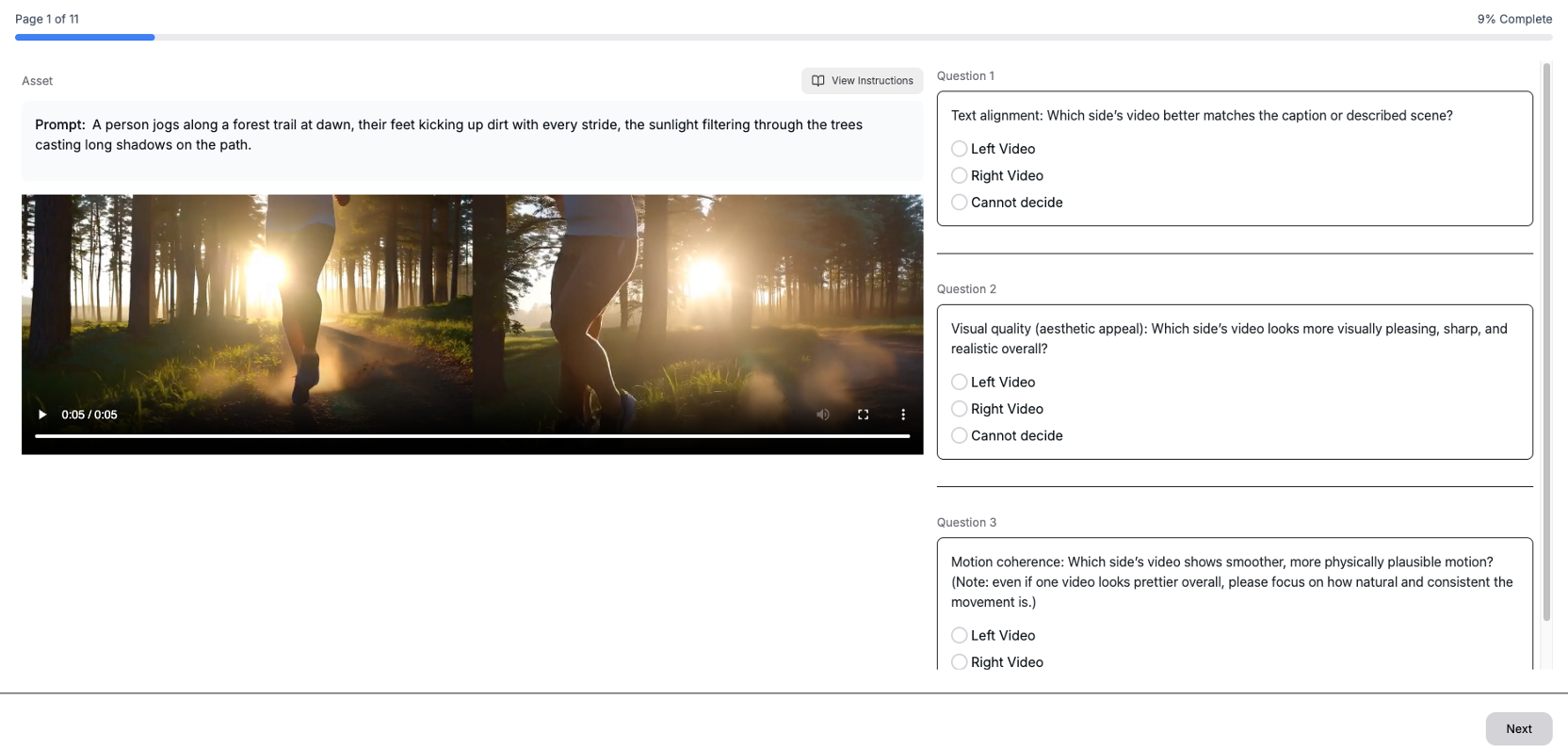}
    \caption{\textbf{Screen-shot of Human Survey Webpage}: three questions regarding visual quality, motion quality and text alignment are asked. The annotators are allowed to answer the question only after they watched the full video.}
    \label{fig:survey-page}
\end{figure}

We provide an screenshot of the huamn survey webpage we created in Figure~\ref{fig:survey-page}.

\end{document}